\documentclass[10pt,twocolumn,letterpaper]{article}

\usepackage{cvpr}
\usepackage{times}
\usepackage{epsfig}
\usepackage{graphicx}
\usepackage{amsmath}
\usepackage{amssymb}

\usepackage{color}
\usepackage{floatrow}
\usepackage{url}
\newcommand\Mark[1]{\textsuperscript#1}

\usepackage[pagebackref=true,breaklinks=true,colorlinks,bookmarks=false]{hyperref}

\cvprfinalcopy 

\def\cvprPaperID{****} 

\ifcvprfinal\fi
\begin{document}

\title{Learning Stylized Character Expressions from Humans}

\author{Deepali Aneja\Mark{1}\qquad Alex Colburn\Mark{2}\qquad  Gary Faigin\Mark{3}\qquad Linda Shapiro\Mark{1}\qquad Barbara Mones\Mark{1}\\
\Mark{1}University of Washington, Seattle\qquad \Mark{2}Zillow Group, Seattle\qquad \Mark{3}Gage Academy of Art, Seattle \\
{\tt\small{\{deepalia,alexco,shapiro,mones\}}@cs.washington.edu, gary@gageacademy.org}
}

\newcommand{\DEEPEXPR}{DeepExpr}
\newcommand{\MTURK}{MT}
\newcommand{\FIGURE}[0]{Fig.}

\maketitle

\section{Introduction}
\vspace{-1.5mm}

We present \textbf{\DEEPEXPR{}}, a novel expression transfer system from humans to multiple stylized characters via deep learning. We developed : 1) a data-driven perceptual model of facial expressions, 2) a novel stylized character data set with cardinal expression annotations : FERG (Facial Expression Research Group) - DB (added two new characters), and 3) a mechanism to accurately retrieve plausible character expressions from human expression queries. We evaluated our method on a set of retrieval tasks on our collected stylized character dataset of expressions. We have also shown that the ranking order predicted by the proposed features is highly correlated with the ranking order provided by a facial expression expert and Mechanical Turk (MT) experiments. 

\section{Related Work}
\vspace{-1.5mm}

Historically, the Facial Action Coding System (FACS) \cite{ekman1978} is often used for Facial Expression Recognition (FER) and character animation systems. We have shown FACS-based systems do not consistently  generate expressions that are recognized by humans as the intended expression. Modern methods apply CNNs to the FER problem and are quite effective in recognition tasks \cite{mollahosseini2016going}. CNN features fused with geometric features for customized expression recognition \cite{yu2016customized} and Deep Belief Networks have also been utilized to solve the FER, but have mixed results when used to generate expressions.

\section{Methodology}
\vspace{-1.5mm}

To learn deep CNN models that generalize well across a wide range of expressions, we need sufficient training data to avoid over-fitting of the model. For human facial expression data collection, we combined publicly available annotated facial expression databases. We also created a novel database (\textbf{FERG-DB}) of labeled facial expressions for six stylized characters. An animator created the key poses for each expression, and they were labeled via \MTURK{} to populate the database initially. The details of the database collection and data pre-processing are given in \cite{aneja2016modeling}. 

\begin{figure}[t]
\centering
\includegraphics[height=1in]{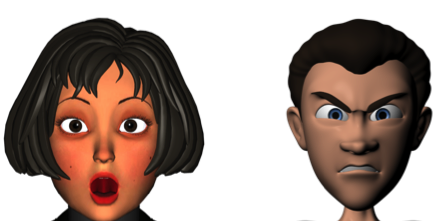}
\caption{Two new characters added to FERG-DB with surprise expression (left) and anger expression (right).}
\label{fig:newCharacters}
\end{figure}

We have extended \textbf{FERG-DB} by adding two new characters as shown in \FIGURE{}~\ref{fig:newCharacters} with almost 10,000 images for each new character. All the images are labeled for each of six cardinal expressions: joy, sadness, anger, surprise, fear, disgust, and neutral.

\subsection{Data Processing and Training}
\vspace{-1.5mm}
For our human dataset, we register faces to an average frontal face via an affine transformation. Geometric measurements between the points are also taken to produce geometric features for refinement of expression retrieval results. Once the faces are cropped and registered, the images are re-sized for analysis.

With approximately 70,000 images of labeled samples of human faces and 50,000 images for stylized character faces, we trained two Convolutional Neural Networks to recognize the expressions of humans and stylized characters independently. Then we utilize a transfer learning technique to learn the mapping from humans to characters by creating a shared embedding feature space. 

To create the shared feature space, we fine-tuned the CNN pre-trained on the human dataset with the character dataset for every character by continuing the backpropagation step. The last fully connected layer of the human trained model was fine-tuned, and earlier layers were kept fixed to avoid overfitting. We decreased the overall learning rate while increasing the learning rate on the newly initialized next-to-last layer. This embedding also allows human expression-based image retrieval and character expression-based image retrieval.

\begin{figure*}[t]
\centering
\includegraphics[width=\textwidth,height=1in]{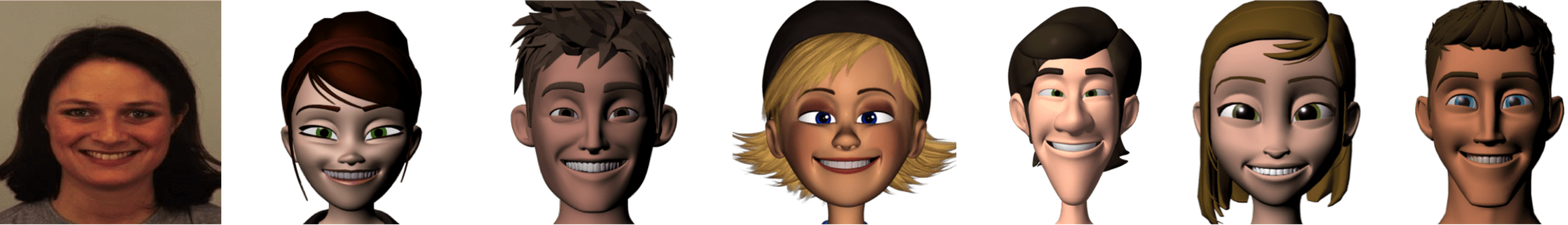}
\caption{Result from our combined approach - \DEEPEXPR{} and Geometric features. The leftmost image is the query image and all six characters are shown portraying the top match of the same joy expression.}
\label{fig:result}
\end{figure*}

\subsection{Distance Metrics}
\vspace{-1.5mm}
In order to retrieve the stylized character with the closest expression match to the human expression, we formulate a distance metric with terms for expression clarity and geometric distance. The closest expression match minimizes the distance function in eq. \ref{eq:opt} :

\vspace{-5mm}
\begin{equation}\label{eq:opt}
\phi_{d} = \alpha \left|{\text{JS Distance}}\right| + 
\beta \left|{\text{Geometric Distance}}\right|
\end{equation}
\vspace{-5mm}

\noindent where the clarity term, JS Distance,  is the Jensen-Shannon divergence \cite{lin1991divergence} distance between next to last layer feature vectors of $human$ and $character$, and Geometric distance is the $L^2$ norm distance between geometric features of $human$ and $character$. Our implementation uses JS Distance as a retrieval parameter and then geometric distance as a sorting parameter to refine the retrieved results with $\alpha$ and $\beta$ as relative weight parameters.

\section{Results and Discussion}
\vspace{-1.5mm}
DeepExpr features combined with geometric features produce significant performance enhancement in retrieving the stylized character facial expressions based on human facial expressions. The top results for a human joy expression on six stylized characters are shown in \FIGURE{}~\ref{fig:result}.

\begin{figure}[b]
\centering
\includegraphics[height=1.5in]{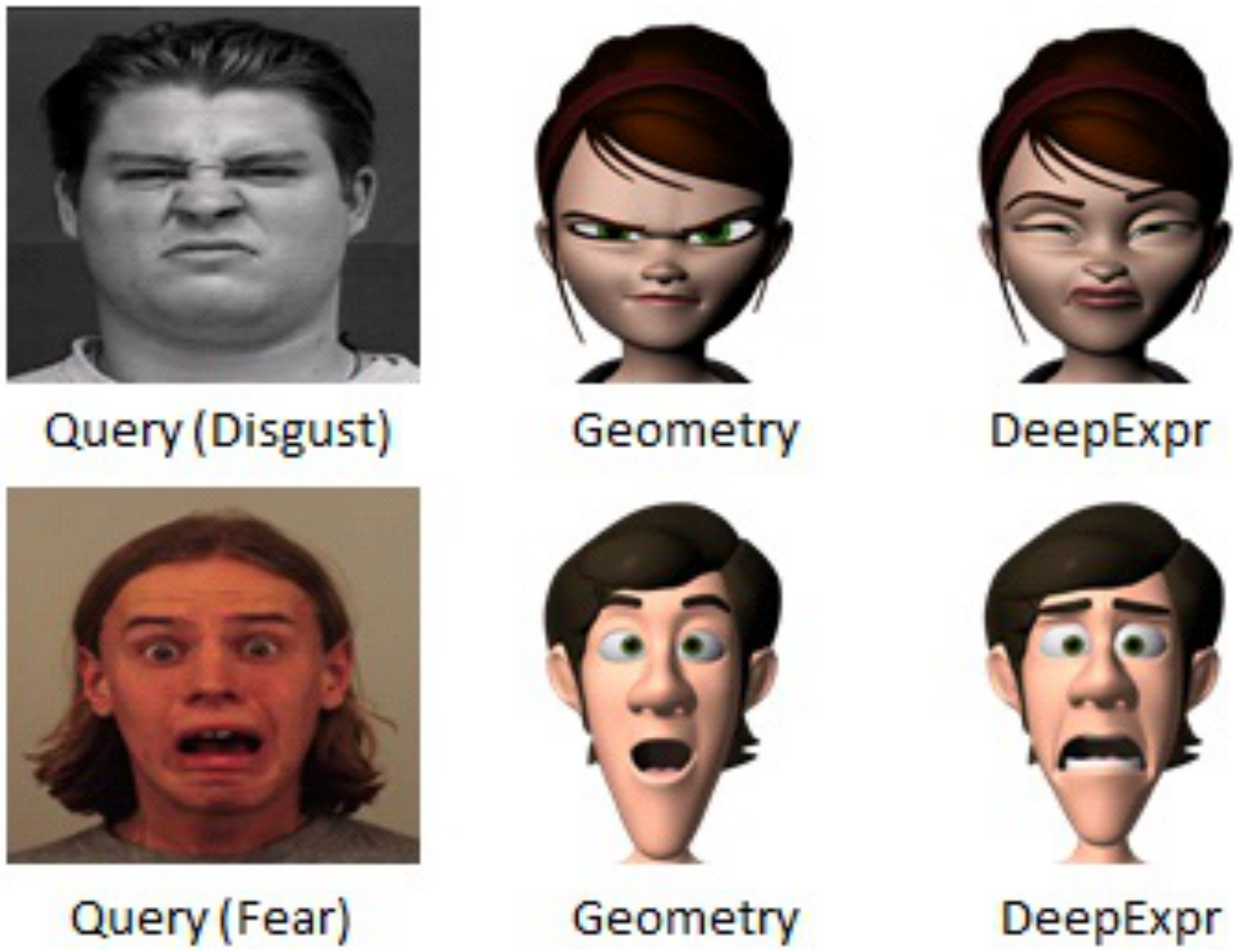}
\caption{Best match results from our approach compared to only geometry-based retrieval for disgust (top) and fear (bottom).}
\label{fig:deepGeo}
\end{figure}

We measured the retrieval performance of our method by calculating the average normalized rank of relevant results \cite{muller2002truth}. The best score of $0$ indicates that all the relevant database images are retrieved before all other images in the database. A score that is greater than 0 denotes that some false positives are retrieved before all relevant images. Table \ref{tab:dist} shows that \DEEPEXPR{} consistently achieves lower scores than geometry alone, and Figure \ref{fig:deepGeo}  illustrates with examples. 

\begin{table}
\begin{center}
 \begin{tabular}{|c |c |c|} 
 \hline
 Expression & Geometry & \DEEPEXPR{} \\
 \hline\hline
 Anger & 0.384 &0.213 \\
 \hline
 Disgust & 0.386 &  0.171\\ 
 \hline
 Fear & 0.419 & 0.228 \\
 \hline
 Joy & 0.276 &0.106 \\
 \hline
 Neutral & 0.429 &0.314 \\
 \hline
 Sad & 0.271 &0.149 \\
 \hline
 Surprise & 0.322 &0.125 \\
 \hline
\end{tabular}
\end{center}
\caption{Average retrieval score for each expression across all characters using only geometry and \DEEPEXPR{}  features.}
\label{tab:dist}
\end{table}

The Spearman and Kendall correlation coefficients of our combined approach ranking with the expert ranking and \MTURK{} test ranking for 30 validation experiments show high correlation score of more than 0.8 for 93\% of the experiments. 

Our system demonstrates a perceptual model of facial expressions that provides insight into facial expressions displayed by stylized characters. The model can also be incorporated into the animation pipeline to help animators and artists to better understand expressions, communicate how to create expressions to others, and transfer expressions from humans to stylized characters.\\

\vspace{-6mm}

{\small
\bibliographystyle{IEEE}
\bibliography{references}
}

\end{document}